\documentclass[conference]{IEEEtran}
\IEEEoverridecommandlockouts
\usepackage{cite}
\usepackage{amsmath,amssymb,amsfonts}
\usepackage{algorithmic}
\usepackage{graphicx}
\usepackage{textcomp}
\usepackage{xcolor}
\usepackage{multirow}
\usepackage{makecell}
\def\BibTeX{{\rm B\kern-.05em{\sc i\kern-.025em b}\kern-.08em
    T\kern-.1667em\lower.7ex\hbox{E}\kern-.125emX}}
\begin{document}

\title{A Multi-Expert Large Language Model Architecture for Verilog Code Generation\\
}

\author{\IEEEauthorblockN{1\textsuperscript{st} Bardia Nadimi}
\IEEEauthorblockA{\textit{Dept. of Computer Science and Engineering} \\
\textit{University of South Florida}\\
Tampa, Florida, United States \\
bnadimi@usf.edu}
\and
\IEEEauthorblockN{2\textsuperscript{nd} Hao Zheng}
\IEEEauthorblockA{\textit{Dept. of Computer Science and Engineering} \\
\textit{University of South Florida}\\
Tampa, Florida, United States \\
haozheng@usf.edu}
}

\maketitle

\begin{abstract}
Recently, there has been a surging interest in using large language models (LLMs) for Verilog code generation.  
However, the existing approaches are limited in terms of the quality of the generated Verilog code.
To address such limitations, this paper introduces an innovative multi-expert LLM architecture for Verilog code generation (MEV-LLM).
Our architecture uniquely integrates multiple LLMs, each specifically fine-tuned with a dataset that is categorized with respect to a distinct level of design complexity. 
It allows more targeted learning, directly addressing the nuances of generating Verilog code for each category. 
Empirical evidence from experiments highlights notable improvements in terms of the percentage of generated Verilog outputs that are syntactically and functionally correct. 
These findings underscore the efficacy of our approach, promising a forward leap in the field of automated hardware design through machine learning.
\end{abstract}

\begin{IEEEkeywords}
Large Language Models, Multiple Expert architectures, fine-tuning, Verilog, dataset, transformers
\end{IEEEkeywords}

\section{Introduction and Motivation}

The driving force behind employing large language models (LLMs) for hardware code generation lies in creating a tool that eases the hardware modeling process for designers. 
The integration of LLMs into this domain is aimed at simplifying the complexities typically associated with hardware modeling. 
By leveraging the advanced capabilities of these models, we can expect that the development of hardware models will become more efficient and intuitive.
Furthermore, the automation aspect of hardware code generation using LLMs plays a crucial role in minimizing human errors. 
Manual coding, especially in the intricate field of hardware designs, is often susceptible to mistakes due to the detailed and technical nature of the work \cite{Hardfails}. 
By automating this process through LLMs, the tool not only accelerates the development cycle but also significantly reduces the likelihood of errors that human designers might introduce. 
This leads to more robust and reliable hardware designs, as the automated system can consistently generate high-quality code with a lower risk of faults. In essence, the use of LLMs in this context represents a step towards more efficient, error-resistant, and user-friendly hardware design methodologies.

The debut of attention-based models~\cite{attentionIsAllYouNeed} marks a pivotal moment in the language processing field. 
This leads to the widespread adoption of transformer architectures by researchers, fueling significant progress in the field.
Numerous models, including Generative Pre-trained Transformers~\cite{radford2018GPT}, Bidirectional Encoder Representations from Transformers~\cite{BERT}, and Language Model for Dialogue Applications~\cite{LaMDA}, utilize transformer technology to achieve good results. 
While there is a significant amount of research dedicated to the synthesis of software programs, the field of hardware code synthesis has seen relatively less exploration. 
Lately, several researchers have investigated using current LLM architectures for generating code in Hardware Description Languages \cite{BenchmarkingVerilog, VerilogEval, ChipGPT, RTLLM, ChipChat, GPT4AIGchip, ChipNeMo, RTLFixer, AutoChip, rtlcoder, improvingLLMforHDL} and in other hardware related problems \cite{distilBERT}. 
While previous work has pioneered HDL code generation for hardware designs using LLMs, there are challenges that need to be addressed for further enhancement. 
A notable challenge is the limited availability of labeled data for efficient fine-tuning of LLMs as in the case of software code generation. Another issue is the relative scarcity of innovative fine-tuning methodologies in this field, unlike what is seen in software code generation~\cite{softwareLLM}. 
The Verilog code generated by the existing approaches often contains syntax errors and functionality errors~\cite{BenchmarkingVerilog, VerilogEval}.

\begin{figure}
    \centering
    \includegraphics[width=\columnwidth]{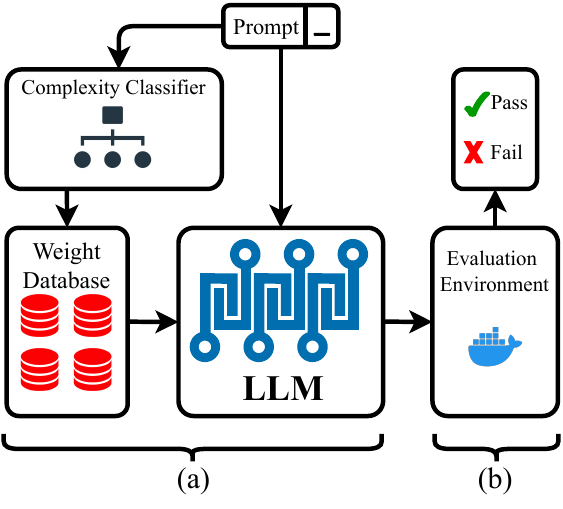}
    \caption{Proposed Overall Architecture. a) MEV-LLM architecture b) Evaluation process}
    \label{fig:MEV-LLM}
\vspace*{-10pt}
\end{figure}

In our study, we introduce a new LLM architecture that utilizes a multi-expert strategy for Verilog code generation based on the complexity level of the design as implied by the input prompt.
The proposed architecture aims to improve the performance of LLMs for generating correct Verilog code by utilizing multiple expert models, each of which is specifically fine-tuned for a category of designs of similar complexity level.
The proposed multi-expert LLM architecture is depicted in Fig.~\ref{fig:MEV-LLM}. 
Moreover, to facilitate the use of the suggested architecture, we develop a diverse and categorized dataset where designs are classified into categories by their complexities, thereby improving the effectiveness of LLM fine-tuning. 
The \textbf{key contributions} of this paper are outlined as follows:

\begin{itemize}
    \item Introduction of a Multi-Expert LLM architectural framework (MEV-LLM), where each expert is fine-tuned with a distinct category of hardware designs of similar complexity.  This architecture ensures that these expert models are finely tuned to the specific requirements of their respective categories.
    \item Development of a dataset categorized by different design complexities for fine-tuning the proposed MEV-LLM architecture.
    Furthermore, additional descriptive details are associated with each entry in the dataset to facilitate a more refined fine-tuning process.
    \item Experiments show that the new multi-expert approach improves Verilog code generations by up to $23.9\%$ using $pass@k$ metric compared with the state-of-the-art approaches including CodeGen-Verilog \cite{codegen, BenchmarkingVerilog} and GEMMA \cite{gemma} architectures.
\end{itemize}

The subsequent sections of this paper are structured in the following manner: Section II provides an overview of prior research in this area. The methodology and the proposed dataset are introduced in section III. Section IV is dedicated to comparisons and evaluations, respectively. Finally, we present our conclusions and future works in Section V.
\vspace*{-6pt}
\section{Related Works}

The study in \cite{BenchmarkingVerilog} examines how LLMs can generate Verilog code, essential for hardware design, achieving notable success in producing syntactically accurate code by fine-tuning them on Verilog-specific datasets. Highlighting LLMs' role in minimizing human errors and advancing automation in hardware design, the research identifies the potential for further improvements in functional correctness and lays the groundwork for future AI integration in hardware development.

The work in~\cite{VerilogEval} introduces VerilogEval, a new benchmarking framework designed to test the effectiveness of LLMs in generating Verilog code for hardware design and verification. Using a dataset of 156 problems from HDLBits, it explores the automation of Verilog code generation for a range of complexities, from basic circuits to intricate finite-state machines. The study shows that fine-tuning LLMs enhances their code generation quality, underscoring the significant role of AI in refining hardware design workflows. It also discusses the potential of supervised fine-tuning to boost LLM performance, marking a valuable advancement for both academic and practical applications in the field.

The study in~\cite{ChipGPT} explores the use of LLMs, such as ChatGPT, for hardware logic design, aiming to translate natural language into hardware logic without extensive human input. It introduces ChipGPT, a novel, scalable framework that simplifies logic design through LLMs, requiring no retraining. The process involves generating Verilog code from LLM prompts, refining the output, and selecting the best design based on specific metrics. ChipGPT enhances programmability and controllability, expanding the possibilities for design optimization beyond traditional methods and existing LLM capabilities.

The work in~\cite{RTLLM} tackles the lack of standardized benchmarks in evaluating LLMs, like ChatGPT, for hardware design, focusing on Register Transfer Level (RTL) creation from natural language. Introducing RTLLM, an open-source benchmark aimed at assessing LLMs' effectiveness in RTL design, the study sets new standards across syntax, functionality, and design quality. It also unveils a unique prompt engineering method, "self-planning," which boosts GPT-3.5's performance, marking a significant advance in utilizing LLMs for complex and scalable hardware design tasks.

The study in~\cite{ChipChat} explores the fusion of conversational AI with hardware design, focusing on automating the translation of natural language specifications into HDLs like Verilog, a task traditionally performed by humans. Utilizing advanced LLMs such as OpenAI's ChatGPT and Google's Bard, the study presents a groundbreaking case study, 'Chip-Chat', where an engineer and an LLM co-design an 8-bit microprocessor, resulting in the first entirely AI-generated HDL for tape-out. This marks a significant advance in integrating AI into hardware design processes.

Each of the previously discussed studies faces challenges in producing Verilog codes that are both syntactically and functionally accurate.
\section{Methodology and Dataset}

This section begins with an explanation of the multi-expert LLM architecture, followed by detailing the process of the categorized dataset. 

\subsection{Multi-Expert Verilog LLM (MEV-LLM)}
One of the limitations of prior research is the approach to fine-tuning. 
Sophisticated LLM-based software code generators typically employ more refined fine-tuning techniques and architectures~\cite{softwareLLM}. 
Earlier studies have opted to refine a single-expert LLM using an entire dataset. 
Verilog coding encompasses various styles, such as combinational or sequential coding. Additionally, the coding strategy may vary depending on the design's complexity. 
With only one expert to manage various design types across multiple complexity levels, the capacity to distinguish between these levels of complexity is restricted.

Contrarily, in this study, we advocate for the employment of multiple expert LLMs, each fine-tuned with datasets tailored to distinct design complexity levels.
The diagram of the proposed architecture is depicted in Fig.~\ref{fig:MEV-LLM}(a). By employing multiple experts, each tailored to a specific level of design complexity, the ability to discern between complexities and apply the appropriate coding style becomes unrestricted. Consequently, this leads to a more accurately selected method for generating the relevant code.
Our proposed methodology encompasses five distinct fine-tuned LLMs. 
In the first stage, we deploy a complexity classifier LLM. 
This classifier model is fine-tuned with a set of design descriptions and their associated complexity categories. 
This set of design descriptions spans all design complexity levels.
The goal of this model is to first assess the complexity level of a design described in the input prompt, and then help to select an expert model for the target category.  

In the second phase, our architecture is equipped with four distinct sets of fine-tuned weights for four expert models fine-tuned with four different categories of a dataset.
Each expert model is tailored to generate Verilog code of a specific level of design complexity. 
Once the complexity classifier model successfully identifies the complexity level of the desired design, the appropriate set of weights is then loaded into the LLM's architecture at this stage. Thus, when activated with the input prompt, the model is capable of producing outputs that are specifically tailored to the determined complexity level, utilizing the weights pertinent to that specific complexity tier.

\subsection{Dataset Development Method}
A high-quality dataset plays a pivotal role in the effective fine-tuning of LLMs. 
The addition of descriptive annotations for each entry in the dataset can markedly refine the fine-tuning process \cite{datasetQuality}, ensuring a closer match between the input data and its associated descriptions, which aids in optimizing the model. 
Previous studies have shown that even partially labeled datasets can enhance the fine-tuning process \cite{semi-labeled}. 
Existing datasets for Verilog codes often suffer from partial labeling, contain a high degree of duplicate entries, and lack categorization. 
In our research, we compile a dataset by extracting Verilog files from public GitHub repositories and crafting descriptions for them based on their readme files and the code content itself, utilizing the chatGPT-3.5-Turbo API for description generation. 
This process of incorporating descriptions into the dataset is referred to as {\em fine-grained labeling}.

As previously outlined, the dataset we propose is organized according to the complexities of the designs to support the proposed multi-expert architecture. 
To achieve this, we identify four distinct categories of design complexity: 
\begin{itemize}
    \item [1)] {\tt Basic}, encompassing straightforward Verilog codes like wirings and elementary gate designs;
    \item [2)] {\tt Intermediate}, covering uncomplicated components such as multiplexers, adders, and arithmetic units;
    \item [3)] {\tt Advanced}, featuring more complex designs like sequential circuits and finite state machines;
    \item [4)] {\tt Expert}, typically consisting of composite designs that integrate elements from the earlier categories.
\end{itemize}
To carry out the categorization process, we prompt the chatGPT-3.5-Turbo API again to classify the dataset across four distinct complexity levels. 
For each dataset entry, we begin by detailing the complexities and their definitions in the prompt, followed by adding the synthetic description and the actual Verilog code to the chatGPT API prompt.
This categorization is performed using the generated description in the fine-grained labeling step. We call categorizing the dataset {\em coarse-grained labeling}.

During our use of chatGPT-3.5-Turbo for generating descriptions and categorizing them, we set a token limit of 4096 for each prompt. Upon reviewing all the data collected from GitHub repositories, we accumulated a total of 31,104 data entries, each annotated with both {\em fine-} and {\em coarse-grained labels}.
\vspace*{-6pt}
\section{Evaluation and Discussion}

\subsection{Baseline LLMs and fine-tuning}

In our study, we employ five different LLMs as the basis for our fine-tuning process: CodeGen 2B, 6B, 16B~\cite{codegen}, and GEMMA 2B and 7B~\cite{gemma}. To fine-tune the GEMMA 2B, 7B, and CodeGen 2B and 6B models, we utilize two Nvidia A100 GPUs. However, the substantial size of the CodeGen 16B model necessitates the use of three Nvidia A100 and two Nvidia A40 GPUs, providing a combined total of 336 GB of GPU memory. The selection of these foundational models is based on evidence from prior research indicating that the CodeGen pre-trained model outperforms its alternatives. Additionally, we opted for GEMMA to facilitate a comparative analysis between CodeGen and a model that has not yet been thoroughly investigated. The baseline LLM architectures selected for this study, along with the fine-tuning parameters employed, are concisely documented in Table.~\ref{tab:fine-tuning_data}.

\subsubsection{Fine-tuning MEV}
For each of these existing models, four expert models are fine-tuned, each of which is tailored for a distinct complexity class (basic, intermediate, advanced, and expert) based on both {\em fine-grained} and {\em coarse-grained} labeled datasets. 
We utilize coarse-grain labels to filter data entries for a specific category and employ fine-grain labels during the fine-tuning process as description and code pairs. Furthermore, another pre-trained model is fine-tuned using a dataset comprised of different problem description and their corresponding design complexity to perform complexity classification at the initial stage of our architecture.

\subsubsection{Supervised Fine-tuning}
To evaluate the impact of fine-grain labels, we additionally fine-tune a single expert model for the baseline models using only {\em fine-grained} labels. During this phase, the models go under fine-tuning using pairs of descriptions and codes encompassing all categories present in the dataset, without any categorial division.

\begin{table}
    \centering
    \caption{Pre-trained LLM architectures and fine-tuning times}
    \label{tab:fine-tuning_data}
    \resizebox{\columnwidth}{!}{%
    \begin{tabular}{c|cccc|c|c} 
         Model       & Layers & \# of Heads & Head Size & Context Size & learning rate & \# of epochs  \\ \hline
         CodeGen-2B  & 32     & 32         & 80         & 2048         &  \multirow{3}{*}{$5e^{-5}$} &  \multirow{3}{*}{$1,5,10$} \\ 
         CodeGen-6B  & 33     & 16         & 256        & 2048         &  &   \\ 
         CodeGen-16B & 34     & 24         & 256        & 2048         &  &   \\ \hline
         GEMMA-2B    & 18     & 8          & 256        & 8192         &  \multirow{2}{*}{$2e^{-4}$} &  \multirow{2}{*}{$1,5,10,20$} \\ 
         GEMMA-7B    & 28     & 16         & 256        & 8192         &  &  \\ 
    \end{tabular}%
    }
\vspace*{-10pt}
\end{table}

\subsection{Evaluation, Comparison, and Explanation of the Results}
The output of the Fig.\ref{fig:MEV-LLM}(a) goes into the evaluation environment in Fig.\ref{fig:MEV-LLM}(b). 
In our analysis, we utilize the evaluation methodology proposed by Liu et al.~\cite{VerilogEval}, leveraging a collection of problems sourced from the HDLBits website. 
As some tasks containing non-textual elements like diagrams and schematics, they are classified into two distinct groups: one where non-text elements were manually translated into text, known as Verilog-Human, with 156 problems, and another where chatGPT-3.5-Turbo is utilized to transform non-text elements into text, labeled as Verilog-Machine, comprising 143 problems.
In their assessment framework, they employ the $pass@k$ metric, signifying that a problem is deemed resolved if at least one out of the k-generated samples successfully passes the unit test.

For the purpose of benchmarking our proposed LLM architecture against prior models, we supply both the existing models and our novel model with identical input prompts. These prompts include a preliminary section requesting Verilog code corresponding to a subsequent description. This description is derived from one of two designated sets (Verilog-Machine or Verilog-Human). For every problem, we produce 15 unique results to evaluate the metrics $pass@1$, $pass@5$, and $pass@10$.

For the CodeGen model, we perform two types of fine-tuning: one involves training the model on our dataset as a whole without any categorization (supervised fine-tuning), and the other involves separately fine-tuning distinct models for each category within our dataset. We then compared these outcomes with the results from the model fine-tuned by Thakur et al. \cite{BenchmarkingVerilog}.
For the GEMMA model, due to the absence of a pre-existing model specifically fine-tuned for Verilog code generation, we undertook the task of fine-tuning this model using the dataset compiled by Thakur et al. \cite{BenchmarkingVerilog}. Additionally, we fine-tune GEMMA with our own dataset, applying both categorized and uncategorized approaches, mirroring the methodology we used for CodeGen. Tables~\ref{tab:codeGenBestResults} to ~\ref{tab:gemmaBestResults} showcases the best results obtained from various fine-tuning sessions conducted with differing parameters.

\begin{table}
    \centering
    \caption{codegen-verilog vs our proposed models}
    \newcolumntype{?}{!{\vrule width 2pt}}
    \label{tab:codeGenBestResults}
    \resizebox{\columnwidth}{!}{%
    \begin{tabular}{c?c|c|c?c|c|c} 
        \multirow{2}{*}{Model}                       & \multicolumn{3}{|c?}{Verilog-Machine} & \multicolumn{3}{c}{Verilog-Human} \\ \cline{2-7} 
                                                     & $pass@1$ & $pass@5$ & $pass@10$ & $pass@1$ & $pass@5$ & $pass@10$ \\ \Xhline{5\arrayrulewidth}
        CodeGenVerilog-2B\cite{BenchmarkingVerilog}  & 20.1     & 45.4     & 55.2      & 17.9     & 35.8     & 40.3      \\ \hline 
        CodeGen-2B fine-grained labels               & 27       & 53.8     & 58.7      & 21.1     & 41       & 45.5      \\ \hline
        CodeGen-2B MEV-LLM                           & \textbf{44}       & \textbf{60.1}     & \textbf{63.6}      & \textbf{34.6}     & \textbf{51.3}     & \textbf{53.2}       \\ \Xhline{5\arrayrulewidth}
        CodeGenVerilog-6B\cite{BenchmarkingVerilog}  & 35.6     & 51.7     & 55.9      & 23       & 39.7     & 49.3      \\ \hline
        CodeGen-6B fine-grained labels               & 42.6     & 55.2     & 59.4      & 26.9     & 42.9     & 51.9      \\ \hline
        CodeGen-6B MEV-LLM                           & \textbf{57.3}     & \textbf{61.5}     & \textbf{66.4}      & \textbf{42.9}     & \textbf{48}       & \textbf{54.4}       \\ \Xhline{5\arrayrulewidth}
        CodeGenVerilog-16B\cite{BenchmarkingVerilog} & 37.7     & 54.5     & 58        & 28.2     & 45.5     & 51.2      \\ \hline
        CodeGen-16B fine-grained labels              & 47.5     & 58       & 60.8      & 32.7     & 50       & 53.2      \\ \hline
        CodeGen-16B MEV-LLM                          & \textbf{60.8}     & \textbf{63.6}     & \textbf{69.2}      & \textbf{47.4}     & \textbf{55.7}     & \textbf{60.2}       \\
    \end{tabular}%
    }
\vspace*{-10pt}
\end{table}

\begin{table}
    \centering
    \caption{Gemma architecture vs our proposed models}
    \newcolumntype{?}{!{\vrule width 2pt}}
    \label{tab:gemmaBestResults}
    \resizebox{\columnwidth}{!}{%
    \begin{tabular}{c?c|c|c?c|c|c} 
        \multirow{2}{*}{Model}          & \multicolumn{3}{|c?}{Verilog-Machine} & \multicolumn{3}{c}{Verilog-Human} \\ \cline{2-7} 
                                        & $pass@1$ & $pass@5$ & $pass@10$ & $pass@1$ & $pass@5$ & $pass@10$  \\ \Xhline{5\arrayrulewidth}
        GEMMA-2B \cite{gemma}           & 12.5     & 23.7     & 29.3      & 11.5     & 21.7     & 26.9       \\ \hline 
        GEMMA-2B fine-grained labels    & 15.3     & 28.6     & 36.3      & 13.4     & 25.6     & 32.6       \\ \hline
        GEMMA-2B MEV-LLM                & \textbf{31.4}     & \textbf{51.7}     & \textbf{57.3}      & \textbf{27.5}     & \textbf{47.4}     & \textbf{52.5}       \\ \Xhline{5\arrayrulewidth}
        GEMMA-7B \cite{gemma}           & 18.8     & 29.3     & 36.3      & 17.3     & 26.2     & 32.6       \\ \hline
        GEMMA-7B fine-grained labels    & 22.3     & 34.9     & 42.6      & 19.8     & 31.4     & 37.8       \\ \hline
        GEMMA-7B MEV-LLM                & \textbf{41.9}     & \textbf{55.9}     & \textbf{59.4}      & \textbf{37.8}     & \textbf{50}       & \textbf{54.4}       \\ 
    \end{tabular}%
    }
\vspace*{-10pt}
\end{table}

Based on the data presented in Tables~\ref{tab:codeGenBestResults} and ~\ref{tab:gemmaBestResults}, it is evident that supervised fine-tuning offers advantages, leading to improved code generation over unsupervised fine-tuning methods. Additionally, these tables highlight the effectiveness of the multi-expert approach we proposed, demonstrating that employing several models, each tailored to a specific task, enhances the quality of code generation.

First, our fine-grained labels register an increase of up to $9.8\%$ in the $pass@k$ metric compared to the earlier research conducted by \cite{BenchmarkingVerilog}. It's crucial to highlight that the LLM architecture used in both instances is the same, with the distinction lying solely in the training dataset. Unlike Thakur et. al. \cite{BenchmarkingVerilog}, our dataset is enriched with fine-grained labels for the data entries.
Additionally, the MEV-LLM architecture we've introduced shows an improvement of up to $23.9\%$ in the $pass@k$ metric over the previous studies by Thakur et. al. \cite{BenchmarkingVerilog}. This outcome serves as evidence supporting the benefits of the architecture we propose.
\vspace*{-6pt}

\subsection{Dataset Quality}

As highlighted in \cite{VerilogEval}, the integrity of the compiled dataset significantly influences the effectiveness of the supervised fine-tuning procedure. 
Due to limitations in verifying each description added to the dataset via chatGPT-3.5-Turbo, we decide to follow the methodology used by the researchers in \cite{VerilogEval} to verify the labels generated by chatGPT. 
In order to assess the dataset's quality, we randomly shuffle the codes and descriptions among the data entries, creating mismatched sets of codes and descriptions. 
We then proceed to fine-tune the model with this deliberately distorted dataset. 
Table~\ref{tab:datasetQuality} presents the outcomes of fine-tuning the GEMMA-2B model using the flawed dataset alongside the accurate dataset. 
The results clearly demonstrate that the model's code generation capabilities are significantly affected by the erroneous dataset, verifying the correctness of the labels and highlighting the crucial significance of the dataset's quality in the process of fine-tuning.
\vspace*{-7pt}

\begin{table}
    \centering
    \caption{Results for erroneous dataset}
    \newcolumntype{?}{!{\vrule width 2pt}}
    \label{tab:datasetQuality}
    \resizebox{\columnwidth}{!}{%
    \begin{tabular}{c?c|c|c?c|c|c} 
        \multirow{2}{*}{Model} & \multicolumn{3}{|c?}{Verilog-Machine} & \multicolumn{3}{c}{Verilog-Human} \\ \cline{2-7} 
                               & $pass@1$ & $pass@5$ & $pass@10$ & $pass@1$ & $pass@5$ & $pass@10$  \\ \Xhline{5\arrayrulewidth}
        erroneous dataset      & 5.5      & 9.7      & 11.8      & 5.1      & 9.6      & 11.5       \\ \hline
        correct dataset        & 15.3     & 28.6     & 36.3      & 13.4     & 25.6     & 32.6       \\ 
    \end{tabular}%
    }
\vspace*{-10pt}
\end{table}

\section{Conclusion and Future Works}

In this study, we introduce a new multi-expert LLM framework, termed MEV-LLM, along with a categorized dataset enriched with descriptions for each entry. The aim of this new architecture is to overcome the shortcomings of previous studies by facilitating enhanced results across all Verilog varieties and complexity tiers by introducing the multi-expert LLM architecture. The assessment outcomes demonstrate that the MEV-LLM architecture successfully realizes its anticipated benefits, with an improvement of up to $23.9\%$ in the $pass@k$ metric being recorded. We contend that significant potential for further advancements exists, both through the compilation of more varied and comprehensive datasets and through further investigation into various complexity levels.

\end{document}